\documentclass[sigconf]{acmart}

\AtBeginDocument{%
  }

\usepackage{multirow}
\usepackage{booktabs}
\usepackage[ruled,linesnumbered]{algorithm2e}
\usepackage[normalem]{ulem}
\usepackage{balance}
\usepackage{threeparttable}
\usepackage{url}

\usepackage{graphicx}
\usepackage{textcomp}
\usepackage{xcolor}
\usepackage{hyperref}
\usepackage{subcaption}

\copyrightyear{2026}
\acmYear{2026}
\setcopyright{cc}
\setcctype{by}
\acmConference[DAC '26]{63rd ACM/IEEE Design Automation Conference}{July 26--29, 2026}{Long Beach, CA, USA}
\acmBooktitle{63rd ACM/IEEE Design Automation Conference (DAC '26), July 26--29, 2026, Long Beach, CA, USA}
\acmDOI{10.1145/3770743.3803902}
\acmISBN{979-8-4007-2254-7/2026/07}

\begin{document}

\title{Beyond Flat Netlist: Hierarchical Graph Representation Learning for Scalable Analysis of Sequential Circuits}

\author{
Jingyi Zhou$^{1,2}$, Zhengyuan Shi$^{1,*}$, Jiaying Zhu$^{1}$, Ziyang Zheng$^{1}$, Qiang Xu$^{1,*}$
}

\affiliation{
\institution{$^{1}$The Chinese University of Hong Kong, Hong Kong, China}
\city{}\country{}
\institution{$^{2}$Tsinghua University, Beijing, China}
\city{}\country{}
}

\email{zhoujy22@mails.tsinghua.edu.cn, {zyshi21, jyzhu24, zyzheng23, qxu}@cse.cuhk.edu.hk}

\renewcommand{\shortauthors}{Jingyi Zhou et al.}




\begin{abstract}

Circuit Representation Learning (CRL) offers a powerful paradigm to guide and optimize core Electronic Design Automation (EDA) tasks, but its practical adoption is hindered by the immense scale of industrial netlists and a failure to explicitly model register-level temporal dynamics. To overcome these barriers, we introduce DeepSeq3, a novel hierarchical framework that abstracts circuits into a two-level representation: fine-grained combinational subgraphs partitioned by flip-flops (FFs), and a high-level Super-Node Graph (SNG) that models the register-transfer structure. A dual Graph Neural Network (GNN) architecture learns representations at both levels, capturing local Boolean logic and global state transitions. Crucially, we introduce a state-centric pre-training scheme that predicts the reachability between FF states, endowing the model with a deep understanding of temporal behavior. Demonstrated on large-scale benchmarks, DeepSeq3's approach yields superior scalability and richer representations, reducing bounded model checking (BMC) solving time by 18\% while guaranteeing correctness. 


\end{abstract}

\maketitle


\setcounter{footnote}{0}
\begingroup
\renewcommand{\thefootnote}{}
\footnote{Our code is available at \url{https://github.com/cure-lab/DeepSeq3}.}
\renewcommand{\thefootnote}{*}
\footnotetext{Corresponding authors}
\endgroup

\section{Introduction} \label{Sec:Intro}
Learning general circuit representations and deploying them across tasks has become an attractive direction in EDA domain. This paradigm (CRL) has progressed from gate-level encoders \cite{li2022deepgate, shi2023deepgate2, shi2024deepgate3, deng2024less, shi2025deepcell} to sequential modeling\cite{khan2023deepseq, Khan2024DeepSeq2, wang2025moss}. To date, these advances show that learning on netlists can drive practical downstream tasks, including power estimation \cite{Khan2024DeepSeq2, fang2025nettag} and Boolean reasoning \cite{wu2023gamora, shi2025logic}. 

Despite recent progress, learning general representations for sequential circuits remains challenging for two reasons: (1) \textbf{Lack of scalability on large circuit}. Real industrial netlists may contain millions or billions of gates and wires. Flat, gate-centric message passing hits memory ceilings\cite{zheng2025deepgate4}, suffers from over-squashing\cite{shi2024deepgate3}, and becomes inefficient on oversized neighborhoods. In practice, this restricts the maximum design scale processable end-to-end, where MOSS\cite{wang2025moss} and DeepSeq Family\cite{khan2023deepseq, Khan2024DeepSeq2} struggle to handle designs exceeding 10K gates. (2) \textbf{Overlook of model register-level behavior}. Most prior models only treat FFs as more nodes in a largely gate-centric graph, mixing state elements and combinational logic within a single undifferentiated substrate. Unfortunately, such framework weakens the inductive bias for register semantics, which represents state operator and the structure of reachable states. Without an explicit register-level abstraction, models tend to emphasize local Boolean structure while under-representing when and how state actually transitions.

To address the above limitations, we introduce \textit{DeepSeq3}, a novel hierarchical circuit representation learning framework designed to capture the register-level semantics for large-scale sequential circuits. Rather than operating on a flattened circuit graph, DeepSeq3 partitions each design along FF (i.e. register) boundaries into combinational logic subgraphs between FFs and the FFs themselves as super-nodes. Each combinational subgraph is first encoded into high-dimensional embeddings by a specialized GNN tailored to combinational reasoning. We then assemble the full design into a hybrid graph (super-node graph, SNG) and encode with another GNN on this graph, whose node features are the produced combinational cones' embeddings. Such register-transfer-like abstraction both compresses the local information within the combinational cones and makes state explicit. To endow the representation with temporal semantics, we introduce state-centric pretraining that predicts reachable FF states of the induced state machine. 

We evaluate DeepSeq3 across a wider range of downstream tasks, including gate-level dynamic power estimation and register-level BMC. Our experimental results demonstrate the improved scalability and richer register-level semantics compared with prior and pioneering approaches. By deploying DeepSeq3 as a plug-in into the BMC engine, the learning-aided solution can significantly reduce the solving time while preserving the correctness. 

We summarize the contribution of this work as follows:

\begin{itemize}
\item We construct a hierarchical representation for sequential netlist with two types of graphs: combinational logic subgraphs and super-node graphs. We also develop DeepSeq3 framework to process the corresponding graphs hierarchically, where two GNNs encode combinational logic subgraphs and super-node graphs (SNG), respectively. 
\item We introduce the register-level pre-train tasks to supervise our model in register-level. The model propagates messages through integral combinational cones, encodes the FF embeddings, and predicts the reachability and probability of successor register-level states. 
\item We deploy DeepSeq3 model into bounded model checking engine. With the ability of capturing register-level semantics of sequential circuits, our model speedups BMC engine by 18\% on average solving time on Hardware Model Checking Competition (HWMCC) benchmarks\cite{hwmcc24, HWMCC24Data,HWMCC20Data}. 
\end{itemize}

\begin{figure*}[!t]
    \centering
    \includegraphics[width=0.95\linewidth]{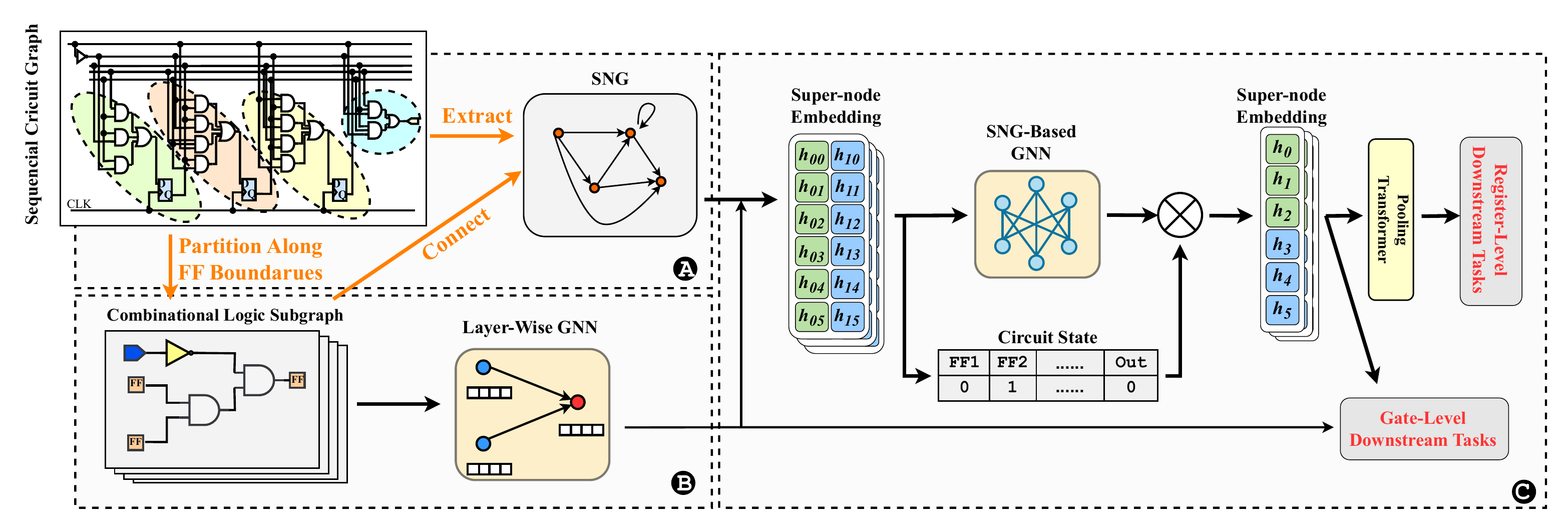} \vspace{-10pt}
    \caption{Overview of DeepSeq3 framework. (A) Subgraph partition and SNG construction. (B) Combinational logic subgraph learning with layer-wise GNN. (C) SNG representation learning framework and applied to various downstream tasks.}
    \label{fig:overview}
    \vspace{-10pt}
\end{figure*}
\section{Related Work}
\vspace{-2pt}
\subsection{Circuit Representation Learning}
\vspace{-2pt}

Circuit Representation Learning (CRL) has made significant progress by capitalizing on the inherently graph-structured nature of netlists. Most existing studies, including Grannite\cite{grannite}, focus on combinational circuits, where logic gates and connections are modeled as Directed Acyclic Graphs (DAGs). By transforming the circuit structure into a graph, CRL enables us to efficiently encode complex circuit information into low-dimensional vectors, thereby greatly empowering downstream EDA tasks.

Extensive work has been done on learning representations for combinational circuits. For instance, DeepGate~\cite{li2022deepgate} introduces attention-based aggregation and skip connections to encode boolean computations among logic gates, improving downstream tasks such as test point insertion~\cite{shi22deeptpi} and Boolean Satisfiability  (SAT)~\cite{Limin23sat}. DeepGate2~\cite{shi2023deepgate2} introduces pairwise truth-table supervision and a functionality-aware loss to improve scalability and efficiency. FGNN~\cite{deng2024less} applies contrastive learning to derive graph-level representations for circuit classification and subgraph localization, while CircuitFusion~\cite{fang2025circuitfusion} and NetTAG~\cite{fang2025nettag} also employ contrastive learning applied to multi-modal circuit data to derive invariant representations. Additionally, Polargate~\cite{liu24polargate} further enhances the circuit representation by employing the bipolar principle. 

Sequential circuit learning remains less explored. DeepSeq~\cite{khan2023deepseq} and DeepSeq2~\cite{Khan2024DeepSeq2} extend CRL to sequential circuits by jointly modeling FFs and combinational logic, learning temporal dependencies through unrolling and sequential message passing. However, while effective, these models primarily operate on bit-level logic and lack explicit graph-level modeling of global circuit structures and connectivity, which limits their ability to capture long-range dependencies. Recently, MOSS~\cite{wang2025moss} leverages a foundation large language model on post-mapping netlists as a general-purpose circuit encoder trained across multiple EDA tasks, enabling cross-task knowledge transfer and improved generalization. A notable limitation of MOSS and the privious DeepSeq family is that they cannot handle large-scale circuits. When applied to extensive netlists, these approoaches suffer from computational overhead and excessive memory consumption.

\subsection{Bounded Model Checking}
\vspace{-2pt}

BMC is a widely adopted formal verification technique used to detect property violations in sequential circuits ~\cite{handbookbmc}. BMC does not construct a complete symbolic model; instead, it unrolls the circuit over a bounded number of time frames and encodes the reachability problem into a SAT instance. This approach offers high efficiency for bug detection but can face prohibitive temporal overhead when applied to large industrial designs. 

Recent studies have explored various methods to accelerate BMC. For example, some approaches handle loops without unrolling by modifying the control flow graph (CFG)~\cite{llm-bmc}, while others intergrated shortcuts into BMC unrolling to accelerate\cite{FrohnGiesl2024arXiv}. Machine-learning-assisted frameworks focus on SAT solving, integrating neural representations to capture the structural or functional properties of circuits~\cite{zhu2025circuit}. 
However, previous works on BMC relies solely on default heuristics and lacks the ability to provide prior knowledge regarding the circuit's reachable states. This fundamental limitation leads to inefficient exploration of the state space.

\section{Methodology} \label{Sec:Method}
\vspace{-2pt}

\subsection{Overview of the Proposed Framework}
\vspace{-2pt}

Figure~\ref{fig:overview} shows the overall operation of the proposed two-stage framework for sequential AND-Inverter Graphs (AIGs). In preprocessing, we decouple combinational logic from sequential elements (FFs) to form multiple independent combinational subgraphs. A layer-wise GNN is then applied within each subgraph to learn node-level representations by propagating information from inputs/predecessors. In the second stage, we abstract the entire circuit into a super-node graph (SNG), where each subgraph acts as one super-node. Another GNN is then employed on this SNG, supervised by both the one-step transition probability matrix and the reachability matrix, to capture structural dependencies and learn the enhanced FF embeddings. The learned embeddings can be used directly for gate-level tasks or aggregated by a Transformer-based pooling module to form the final circuit-level representation.

\subsection{Combinational Logic Subgraph Learning} \label{Sec:Syn}
To derive the representation for each super-node, we employ a tailored GNN on the corresponding partitioned combinatorial logic subgraph for learning. All combinational logic subgraphs are driven by FF outputs and PIs, and they terminate at the next FF or POs. In this manner, each subgraph encapsulates the complete logical signal propagation relationship from the inputs to an output. Since the functions of these subgraphs are time-independent, it serves as the minimal building block for a larger-scale sequential circuit.

We use a layer-wise GNN to learn the functional representation of each combinational subgraph~\cite{shi2023deepgate2}. We denote the logic-1 probability to the likelihood that a node's value is 1, while the logic-0 probability is the probability that its value is 0.
Both logic-0 and logic-1 probabilities are used as supervision signals since they equally determine circuit states~\cite{liu24polargate}. We denote the embeddings learned under these two conditions as $\mathbf{h}^{0}$ and $\mathbf{h}^{1}$. The initial features $\mathbf{h}^{0}_v(0)$ and $\mathbf{h}^{1}_v(0)$ are encoded from the gate type.

For an inverter, the message for one output state comes from the opposite state of the predecessor $u$:
\begin{equation}
    \mathbf{m}_v^0(l) = {aggr}_{NOT}^0(\{\mathbf{h}_u^1(l-1) \mid u \in \mathrm{Predecessors}(v)\})
\end{equation}
\begin{equation}
    \mathbf{m}_v^1(l) = {aggr}_{NOT}^1(\{\mathbf{h}_u^0(l-1) \mid u \in \mathrm{Predecessors}(v)\})
\end{equation}

For an AND gate, the message for an output state aggregates the corresponding state from all predecessors:
\begin{equation}
\mathbf{m}_v^k(l) = {aggr}_{AND}^1(\{\mathbf{h}_u^k(l-1) \mid u \in \mathrm{Predecessors}(v)\}) ,k \in \{0, 1\}
\end{equation}

For any node $v$, its $\mathbf{l}$-th layer state representation $\mathbf{h}_v^k(l)$, is updated using information $\mathbf{m}_v^k(l)$ aggregated from its predecessors:
\begin{equation}
    \mathbf{h}_v^k(l) = \mathrm{GRU}(\mathbf{h}_v^k(l-1), \mathbf{m}_v^k(l)), k \in \{0, 1\}
\end{equation}

\subsection{Hierarchical Graph Representation} 
To learn the global behavior of the register-level behavior, we propose an efficient and robust framework for hierarchical graph learning. 
We first construct the SNG via a proposed super-node partitioning method, which not only significantly reduces the number of nodes but also ensures that all information regarding signal propagation between FFs is preserved.
Next, we define the circuit state and subsequently introduce transition matrices as novel supervision signals for the SNG's representation learning. Finally, we design a gating mechanism and a transformer-based pooling mechanism for the model to capture its temporal semantics.

\subsubsection{SNG Construction Method}
After obtaining embeddings for all combinational subgraphs, we construct the SNG.
The SNG serves as an abstract, register-level graph that captures high-order dependencies among sequential blocks.

\begin{figure}[!t]
   \centering
   \includegraphics[width=0.9\linewidth, height=0.45\linewidth]{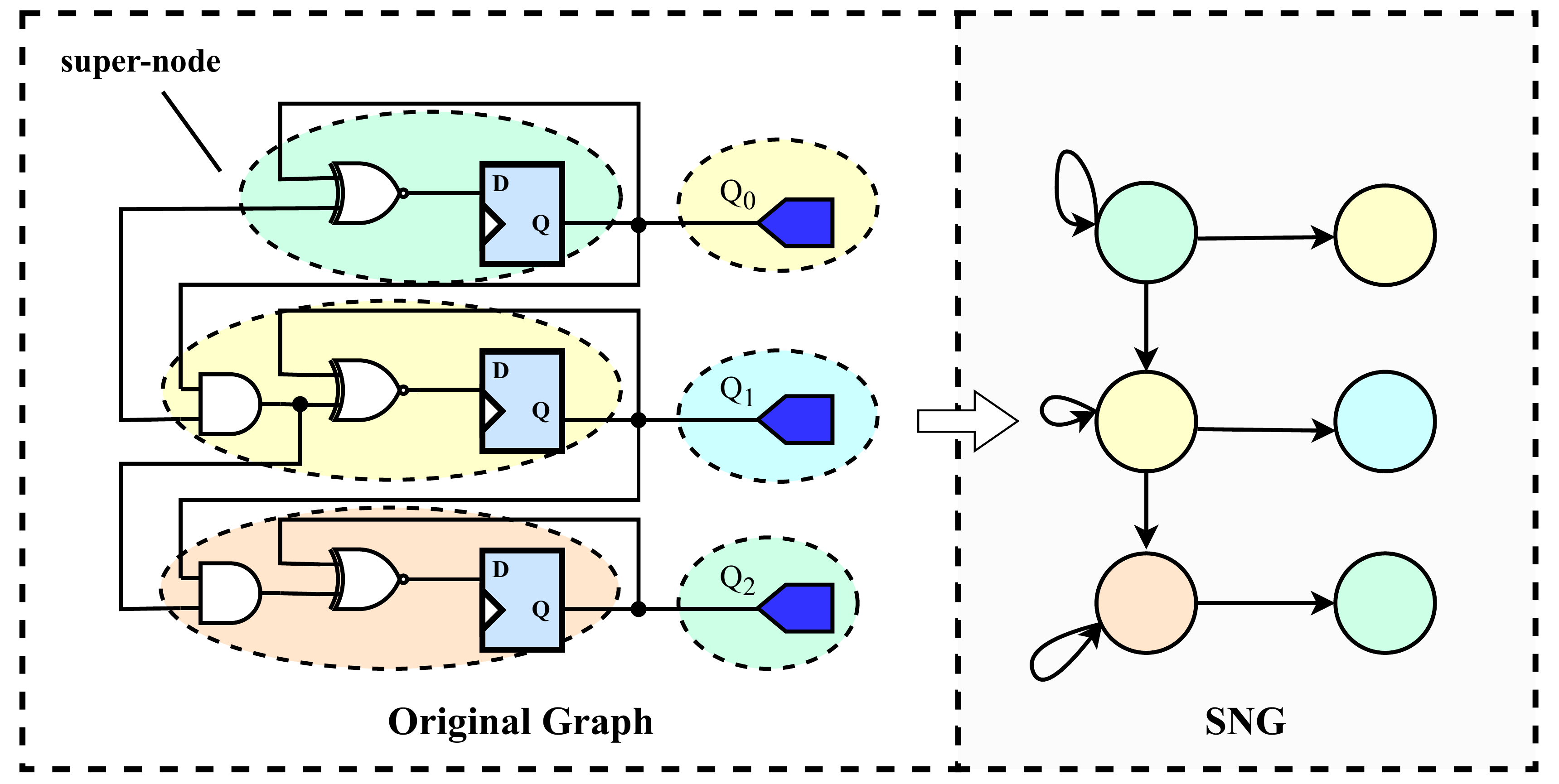} \vspace{-5pt}
   \caption{An example of SNG construction}
   \label{fig:SNG}
   \vspace{-10pt}
\end{figure}

The specific construction method is as follows (Figure ~\ref{fig:SNG}):

\begin{itemize}
    \item Nodes: Each combinational logic subgraph is treated as a super-node, representing a "clock-domain data block" between registers.
    \item Edges: If a logical driving relationship exists between two combination logic subgraphs (i.e., the output FF of subgraph A drives the input FF of subgraph B), a directed edge (super-edge) is added between the corresponding super-nodes.
\end{itemize}

The combinational logic blocks are elevated to node modules, and the sequential flow graph is constructed using FF-to-FF edges. Each super-node represents a localized sequential domain of action and the super-edges collectively form a sequential dependency Directed Graph (DG), which approximates the state propagation map across clock cycles. This graph can then be further encoded using GNNs and RNNs to achieve a higher-level understanding of the circuit's temporal structure.

\subsubsection{Circuit Sequential Modeling}
Previous works, DeepSeq and DeepSeq2, employed different forms of supervision signals to learn the circuit's temporal characteristics~\cite{khan2023deepseq,Khan2024DeepSeq2}. DeepSeq utilized the transition probability of individual FFs as a supervision signal, DeepSeq2 further introduced the state transition similarity between pairs of FFs as supervision to capture the circuit's dynamic correlation. Both methods, however, suffer from a common limitation: they primarily focus on local state relationships and fail to capture the global temporal structure of the circuit. Consider a simple cyclic system with FFs $L_1, L_2,$ and $L_3$:$\begin{cases}
L_1(t+1) = L_2(t) \\
L_2(t+1) = L_3(t) \\
L_3(t+1) = \lnot L_1(t)
\end{cases}$ The circuit can only cycle through a limited set of global states. Supervising only transitions of individual FFs captures local correlations (e.g., $L_1$ follows $L_2$) but fails to reflect the global cyclic dependency and long-term constraints among all FFs.

To address the nuances, we consider its sequential operation as a Finite State Machine. Let the circuit comprise $n$ FFs and $m$ outputs. We formally define the system state vector at time $t$ (e.g., the $t$-th clock rising edge) as:
\begin{equation}
    \mathbf{s}(t) = [q_1(t), q_2(t), \ldots, q_n(t), y_1(t), y_2(t), \ldots, y_m(t)]^{\mathrm{T}}
\end{equation}
where $q_i(t) \in \{0,1\}$ represents the logic value of the $i$-th FF at time $t$, and $y_j(t) \in \{0,1\}$ represents the logic value of the $j$-th output port. Consequently, the complete state space of the circuit is given by $S = \{0,1\}^{n+m}$ with a cardinality of $|S| = 2^{n+m}$.
The temporal evolution of the circuit is formalized as a discrete-time Markov chain. Its state transitions are described by the state transition probability matrix $\mathbf{P} \in \mathbb{R}^{|S| \times |S|}$, where the element $P_{ij}$ is defined as the probability of transitioning from state $s_i$ to state $s_j$:
\begin{equation}
    P_{ij} = \Pr\big(\mathbf{s}(t+1) = s_j \mid \mathbf{s}(t) = s_i \big), \quad \forall s_i, s_j \in S
\end{equation}
It characterizes the state transition distribution driven by the input stimuli and internal logic.

After computing the matrix $\mathbf{P}$, we define the infinite-step reachability matrix $\mathbf{A}_{\infty}$ to describe long-term evolution.:
\begin{equation}
    \mathbf{A}_{\infty}(i,j) =
\begin{cases}
1, & \text{if there exists } k \ge 1 \text{ such that } P^{(k)}_{ij} > 0 \\
0, & \text{otherwise}
\end{cases}
\end{equation}
The matrix $\mathbf{A}_{\infty}$ describes whether state $s_j$ is reachable from state $s_i$ within any number of time steps. Formally:
$$\mathbf{A}_{\infty}(i,j) = 1 \iff s_j \text{ is reachable from } s_i$$

The matrix $\mathbf{P}$ and $\mathbf{A}_{\infty}$ jointly characterize the circuit’s temporal dynamics. Incorporating both as supervision signals enhances the model’s understanding of temporal evolution, enabling it to learn not only local transitions but also the global organization of the circuit’s dynamic behavior.
\vspace{-2pt}

\subsubsection{Ground-Truth Generation}

We employ a simulation-based approach for fast approximation on matrix $\mathbf{P}$. For each state transition $S_i \to S_j$, multiple simulations are performed to record the frequency of transitions from state $S_i$ to $S_j$. The number of valid input vectors $N_{i \to j}$ is counted and normalized by the total input space size $|\mathbf{S}|$ to compute the transition probability $p_{ij} = \frac{N_{i \to j}}{|\mathbf{S}|}$. For small- and medium-scale circuits, where the number of reachable states is limited, we compute the full transition matrix $\mathbf{P}$ directly to preserve accuracy. To handle large-scale circuits, we avoid enumerating all possible state pairs. Instead, we sample a subset of source states $S_i$ from simulation traces and record only the transitions that actually occur among them.
This yields a sparse transition matrix 
$\mathbf{P}$ that captures the dominant transition patterns while keeping both computation and memory cost manageable.




The matrix $\mathbf{A}_{\infty}$ is derived by applying transitive closure to the matrix $\mathbf{P}$ of a Markov chain. $\mathbf{P}$ is first converted into a Boolean adjacency matrix $\mathbf{A}$, where $\mathbf{A}_{ij} = 1$ if and only if $P_{ij} > 0$. The matrix $\mathbf{A}$ is then initialized to ensure self-reachability by setting all diagonal elements to $1$. Similarly, for small-and medium-scale circuits, we employs the Floyd-Warshall algorithm\cite{floyd, warshall} with $O(N^3)$ computational complexity. The iterative update, $\mathbf{A}[i, :] \leftarrow \mathbf{A}[i, :] \lor \mathbf{A}[k, :]$, ensures that if state $i$ reaches $k$, and $k$ reaches $j$, then $i$ can reach $j$. 
For large-scale circuits, we switch to Breadth-First Search (BFS) to mitigate computational complexity. Starting from the source state $S_i$, a queue is utilized to explore all reachable states, and the value of $\mathbf{A}_{\infty}[i][j]$ is updated accordingly. The total complexity for traversing all $N$ states is $\mathbf{O(N^2 + NE)}$. In large-scale sparse scenarios where $E \ll N^2$, the effective complexity can thus be approximated $\mathbf{O(N^2)}$.

Since real verification also explores only partial reachable FF states rather than the full $2^{n+m}$ space, we naturally sample these practically relevant regions, thus avoiding global state explosion.

\subsubsection{Model Pre-training}
\label{Sec:Experiment}
\begin{table*}[!t]
\vspace{-5pt}
\setlength{\tabcolsep}{2pt}
\scriptsize
\centering
\caption{Performance comparison on model pre-training tasks.}
\label{tab:pretrain_results} \vspace{-5pt}
\resizebox{\linewidth}{!}{
\begin{tabular}{lcccccccccccc}
\toprule
\multirow{4}{*}{{\textbf{Model}}}&
\multicolumn{4}{c}{\textbf{Stage 1: Combinational Subgraph}} & &
\multicolumn{6}{c}{\textbf{Stage 2: Register-Level Evaluation}} \\
\cmidrule(lr){2-5} \cmidrule(lr){7-12}
  & \multicolumn{2}{c}{$\mathcal{L}_1$} &  \multicolumn{2}{c}{$\mathcal{L}_0$} &  &
 \multicolumn{3}{c}{IR} & \multicolumn{2}{c}{OT} & Time (s) \\
\cmidrule(lr){2-3} \cmidrule(lr){4-5} \cmidrule(lr){7-9} \cmidrule(lr){10-11}
 & R$^2$ & MAE & R$^2$ & MAE &  &
 Recall & F1 & Accuracy & R & MAE &  \\
\midrule
GCN\cite{kipf2017semi} & 0.7996 & 0.0785 & 0.7910 & 0.0902 &  &- & - & - & -& - & - \\
GAT \cite{velickovic2018gat}& 0.6073 & 0.1117 & 0.5573 & 0.1156 &  &- &-  & - & - &-  & - \\
GraphSAGE\cite{graphsage} & 0.9552 & 0.0367 & 0.9565 & 0.0403 &  &- & -&- &- & -& -&- \\
HOGA-3 \cite{deng2024less}& 0.6066 & 0.1179 & 0.6038 & 0.1186 &  &- & -& -& -& -& -& \\
DeepSeq2\cite{Khan2024DeepSeq2} &- &- &- &- & & 0.2986 & 0.3048 & 0.6714 & 0.2922 & 0.3571 & 40,486 \\
DeepSeq3 w/o Stage1 &- &- &- &- & & 0.8324 & 0.7752 & 0.9467 & 0.7588 & 0.0072 & - \\
DeepSeq3 w/o OT &- &- &- &- & & 0.8406 & 0.8249 & 0.9478 & 0.6442 & 0.0201 & - \\
DeepSeq3 w/o IR &- &- &- &- & & 0.0274 & 0.0341 & 0.8229 & \textbf{0.8695} & \textbf{0.0048} & - \\
DeepSeq3 & \textbf{0.9869} & \textbf{0.0193} & \textbf{0.9916} & \textbf{0.0168} & & \textbf{0.8503} & \textbf{0.8429} & \textbf{0.9564} & 0.8413 & 0.0049 & \textbf{9,896} (3,845+6,051) \\
\bottomrule
\end{tabular}
} \vspace{-5pt}
\end{table*}
We perform global feature modeling on a SNG to enable the model to capture the circuit's structural dependencies and temporal evolution patterns.
The circuit is represented as $G = (V, E)$, where each super-node $v_i \in V$ represents a combinational logic subgraph, and the edges $E$ represent the temporal connection relationships between FFs. The initial feature vector of each node, $h_i^0$ and $h_i^1$, are derived from the combinational logic subgraph embeddings learned in the previous stage.

At the super-node level, each super-node update form for the l-th round is defined as:
\begin{equation}
h_i^k(l) = \text{aggr}^k(h_i^k(l-1), \{h_j^k(l-1) \mid j \in \mathcal{N}(i)\}), k \in \{0,1\}\\
\end{equation}
where $N(i)$ denotes the set of temporal predecessors of node $i$.

To characterize the influence of the circuit state on temporal behavior, we introduce a gated selection mechanism. During the training process, the model simultaneously maintains two sets of state-dependent node embeddings:
\begin{itemize}\item $h_i^0$: The embedding when the local logical state is 0;\item $h_i^1$: The embedding when the local logical state is 1.\end{itemize}

For a given state vector $s=[s_1,s_2,…,s_N]\in \{0,1\}^N$, the active representation is selected as:
\begin{equation}
h_i = s_i \cdot h_i^1 + (1 - s_i) \cdot h_i^0
\end{equation}

The model computes node embeddings under both initial state $s_i$ and target state $s_j$ and obtained the refined gate-level representation $h_i(s_i)$ and $h_i(s_j)$. 
To capture the features of the state transition from $s_i$ to $s_j$, these two representations are concatenated at the node level and form the combined node representation $h'_i = \mathbf{concat}[h_i(s_i), h_i(s_j)]$. These combined node representations are then aggregated through a Pooling Transformer(PT)~\cite{shi2024deepgate3}, which typically uses a specialized [CLS] token to obtain the global circuit representation $\mathbf{H}$: 
\begin{equation}
\mathbf{H} = \text{PT}(\text{[CLS]}, \{\mathbf{concat}[h_i(s_i), h_i(s_j)] \mid i \in V\})
\end{equation}

This global embedding $\mathbf{H}$ synthesizes the circuit's structural topology and the pattern of state change associated with the transition from $s_i$ to $s_j$.
Next, we map the global embedding to the output space using a Multi-Layer Perceptron (MLP) to predict the one-step state transition probability matrix $\mathbf{P}$ and the reachability matrix $\mathbf{{A}_\infty}$. These optimized embeddings serve as the input features for subsequent EDA tasks.

The update function $\text{aggr}^k$ is instantiated as a structure combining the attention mechanism and Gated Recurrent Unit (GRU).
\begin{equation}
    \hat{h}_v^k(l) = \sum_{u \in \mathcal{N}(v)} \alpha_{uv}^k \cdot h_u^k(l-1) 
\end{equation}
\begin{equation}
    h_v^k(l) = \text{GRU} \left( h_v^k(l-1), \hat{h}_v^k(l) \right)
\end{equation}

\section{Experiments} 
In this section, we elaborate the model pre-training results and evaluate its scalability (see Sec.~\ref{Sec:Exp:Result}). Then, we apply DeepSeq3 framework to both gate-level tasks and register-level tasks: power estimation (Sec.~\ref{Sec:Exp:Power}) and BMC (section~\ref{Sec:Exp:BMC}) to demonstrate its ability.

\subsection{Experimental Settings}
\textbf{\quad Datasets.} We build a dataset of 3,568 sequential circuits from ISCAS’89\cite{Brglez1989ISCAS}, Opencore\cite{Opencores}, and ITC’99\cite{Davidson1999ITC} benchmarks. All circuits are processed into the sequential AIG format using the ABC tool \cite{brayton2010abc,mishchenko2007abc}.
Each circuit contains about 100–500 nodes after optimization.

\textbf{Implementation Details.} All models use a hidden dimension of 128, and a 3-layer pooling transformer for global representation.
For all downstream tasks, the final graph embedding is processed by a 3-layer multilayer perceptron (MLP). 
We use the Adam optimizer with learning rate $1 \times 10^{-4}$ and train 40 epochs.
All calculations and model training are performed with 2 $\times$ $\text{NVIDIA A100}$ $\text{GPUs}$.

\textbf{Evaluation Metrics.}
For the subgraph learning stage (Stage1), we evaluate logic-0 and logic-1 probability prediction using Mean Absolute Error ($\text{MAE}$) and the Coefficient of Determination ($\text{R}^2$ Score). The total loss $\mathcal{L}_{\text{total}}$ is defined as the mean of the logic-1 probability loss ($\mathcal{L}_1$) and the logic-0 probability loss ($\mathcal{L}_0$).

For the SNG stage (Stage2), we evaluate:
\begin{itemize}
    \item Infinite Reachability (IR): a binary classification task, measured by Recall, F1, and Accuracy;
    \item One-step Transition (OT): a regression task, measured by Correlation Coefficient (R) and MAE.
\end{itemize}
\vspace{-3pt}

\subsection{Model Pre-training Results} \label{Sec:Exp:Result} \vspace{-2pt}
This section presents a detailed comparative analysis to validate the temporal semantic capture capabilities of the DeepSeq3 framework.

For Stage1, as our subgraph encoder shares the layer-wise design of the DeepGate family~\cite{shi2023deepgate2}, we focus on generic GNNs to demonstrate the necessity of logic-aware inductive bias. Therefore, we choose  $\text{GCN}$\cite{kipf2017semi}, $\text{GAT}$\cite{velickovic2018gat}, $\text{GraphSAGE}$\cite{graphsage}, and $\text{HOGA-3}$\cite{deng2024less} as baselines.
For Stage2, due to the reason that MOSS\cite{wang2025moss} mainly focuses on post-mapping netlists, we choose the current State-of-the-Art method on AIGs, $\text{DeepSeq2}$, as the main baseline. We also include abation settings: DeepSeq3 without combinational subgraph learning is denoted as \textbf{DeepSeq3 w/o Stage1}, DeepSeq3 without IR supervision is denoted as \textbf{DeepSeq3 w/o IR}, DeepSeq3 without OT supervision is denoted as \textbf{DeepSeq3 w/o OT}.
The results are summarized in Table\ref{tab:pretrain_results}.

\textbf{Performance on Combinational Subgraphs.}
The results show that DeepSeq3's customized GNN demontrates stronger logic reasoning in combinational logic. For logic-1 probability prediction, DeepSeq3 achieves an $R^2$ of $0.9869$ and a MAE of $0.01936$. Similarly, for logic-0 probability, it reaches an $R^2$ of $0.9916$ and a MAE of $0.01687$. In contrast, co general GNN models like GCN and GAT yield $R^2$ values significantly below $0.8$. Notably, the GraphSAGE model achieves a higher $R^2$ of approximately $0.95$ in this task. 

\textbf{Performance on Register-Level Representation.}
Our model outperforms DeepSeq2 in register-level tasks in Stage2.
For the IR task, The F1 of DeepSeq3 is $0.8429$, while DeepSeq2 only achieves $0.3048$. For the OT task, DeepSeq3’s R value reaches $0.8413$, significantly higher than DeepSeq2’s 0.2922. The performance of SNG representation gain confirms that our framework successfully learns a richer and more scalable register-level representation.  

\textbf{Ablation Study.}
To quantify the contribution of each component, we conduct an ablation study. The results reveal the effectiveness of subgraph representations and the joint supervision. DeepSeq3 w/o Stage1 results in a drop of the IR F1 Score to $0.7752$. DeepSeq3 w/o IR achieves the highest $R$ value ($0.8695$) but its F1 Score dropped to $0.0341$. Conversely, DeepSeq3 w/o OT maintains a high F1 Score ($0.8249$) but causes the $R$ value to decline to 0.6442. 

\textbf{Efficiency Comparison.}
Finally, DeepSeq3 improves its efficiency upon the previous model. The total pre-training cost of DeepSeq3 is 9896 seconds compared to DeepSeq2's 40486 seconds, which is an approximate four-fold acceleration. This advantage is attributed to two factors: Stage 1 transforming feature extraction into a fully parallelizable task along FF boundaries, and the SNG construction achieving extreme graph compression by abstracting subgraphs into super-nodes.
\vspace{-3pt}
\subsection{Evaluation on Dynamic Power Prediction}  \label{Sec:Exp:Power}  \vspace{-2pt}

This section evaluates the effectiveness of the DeepSeq3 representations in estimating dynamic power to validate the versatility of the sequential circuit representation.

\textbf{Problem Statement.}The total power consumption of Integrated Circuits is primarily composed of two components: Static Power and Dynamic Power. Dynamic power is caused by the circuit's switching activity, and is indispensable when the chip is operational. Therefore, developing fast and accurate dynamic power prediction techniques during the design phase is a crucial challenge for modern VLSI systems.

\textbf{Dynamic Power Estimation Process.}To ensure the practical utility and generality of the assessment, we employ the Nangate 45nm standard cell library as the technology mapping target and  simulated through 10k clock cycles to determine the ground-true dynamic power. The task involves a MLP layer that maps the SNG node embeddings (i.e., the enhanced FFs' representations) and logic gates' embeddings outputted by DeepSeq3 to the dynamic power prediction.

\textbf{Results.}The result was shown in Table~\ref{tab:power_comparison}, and the power values are reported in milliwatts (mW). 
DeepSeq3 outperforms DeepSeq2 in dynamic power estimation.
For example, DeepSeq3's average MAPE is only $4.58\%$. This result translates to an approximate $38.4\%$ improvement in relative accuracy.
DeepSeq3 also shows better robustness in its results compared to DeepSeq2. 
The standard deviation ($\sigma$) of the error for DeepSeq3 is $6.64\%$, which is significantly lower than DeepSeq2’s $7.49\%$. This accuracy gain is derived from the explicit, global temporal dynamics supervision introduced in DeepSeq3, a feature not explicitly modeled in previous methods.

\begin{table}[!t]
\setlength{\tabcolsep}{2.5pt}
\centering
\caption{Predicted Power Comparison}
\label{tab:power_comparison} \vspace{-5pt}
\resizebox{\columnwidth}{!}{
\begin{tabular}{l c c c c|c c}
\toprule
\multirow{2}{*}{\textbf{Circuit}} & \multirow{2}{*}{\textbf{Nodes}} & \multirow{2}{*}{\textbf{Power}} & \multicolumn{2}{c}{\textbf{DeepSeq2}} &  \multicolumn{2}{c}{\textbf{DeepSeq3}} \\
\cmidrule(lr){4-7} 
& & & \textbf{Power} & \textbf{Error}  & \textbf{Power} & \textbf{Error}  \\
\midrule
ptc & 2,024 & 0.0677 & 0.0710  & \textbf{4.89\%} & 0.0717 & 5.92\% \\
noc\_router & 5,246 & 0.2184 & 0.2288  & 4.76\% & 0.2196 & \textbf{0.55\%} \\
mem\_ctrl1 & 10,733 & 0.3148 & 0.3800  & 20.70\% & 0.3641  & \textbf{15.66\%} \\
ac97\_ctrl & 14,004 & 0.5512 & 0.5391  & 2.24\% & 0.5517 & \textbf{0.09\%} \\
pll & 18,208 & 0.7787 & 0.7428  & 4.61\% & 0.7735 & \textbf{0.67\%} \\
\midrule
$\mathrm{Avg.\pm std.}$  & &  & &7.44\% $\pm$ 7.49\% &  & \textbf{4.58\%} $\pm$ \textbf{6.64\%}\\
\bottomrule
\end{tabular}
} \vspace{-8pt}
\end{table}

\subsection{Evaluation on Bounded Model Checking}  \label{Sec:Exp:BMC}
This section validates the direct applicability of DeepSeq3's learned high-level temporal representation in formal verification. 

\textbf{Problem Statement.}
BMC is a crucial formal verification technique that encodes the circuit's reachability property as a SAT problem. However, current approaches lack effective guidance for optimizing reachability search and generating high-quality initial solving assignments, which severely limits its efficiency for large-scale and deep bug detection.

\textbf{Scalable Fine-Tuning for Large-Scale Circuits.}
Our approach addresses this via a lightweight, per-circuit fine-tuning step.
When targeting a new, large circuit for BMC, we first apply the hierarchical partitioning to generate its SNG. Critically, the SNG's node count is one to two orders of magnitude smaller than the original gate-level graph. This graph compression is the key to scalability: even for a massive 96K-node circuit, its SNG remains manageably small.
We then fine-tune the pre-trained DeepSeq3 model on this specific circuit. The ground truth is generated using sampling-based approximation of the $\mathbf{P}$ and $\mathbf{A_{\infty}}$ matrices (Section 3.3.3).

\begin{figure}[!t]
   \centering
   \includegraphics[width=0.8\linewidth]{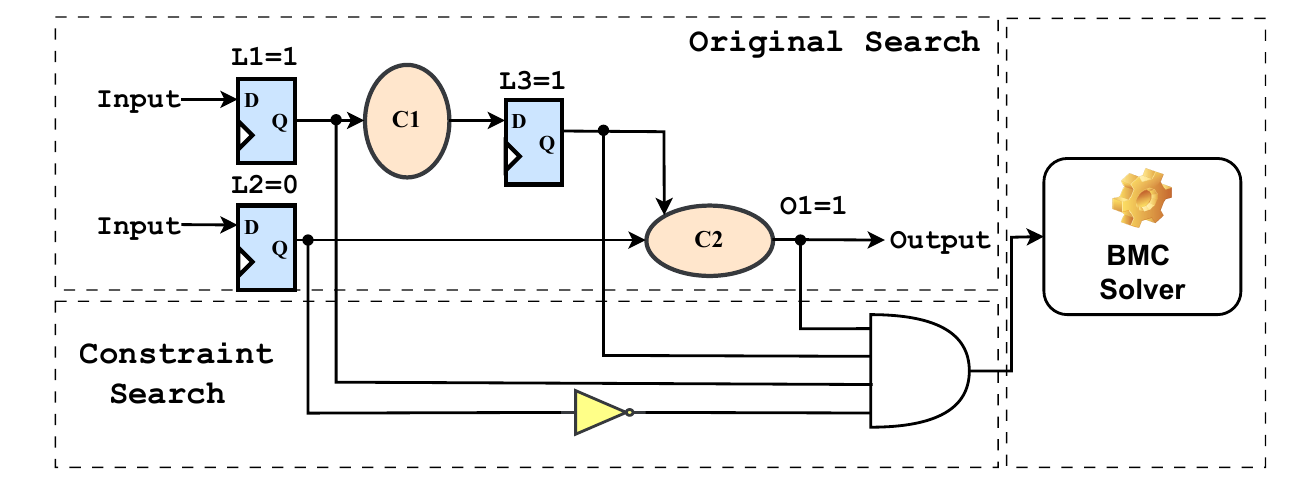} \vspace{-5pt}
   \caption{Guided BMC schematic}
   \label{fig:BMC}
   \vspace{-10pt}
\end{figure}

\textbf{Guided BMC Process.} After fine-tuning, we integrate DeepSeq3 representations to guide the search process of an incremental BMC solver. In the standard flow, the SAT solver performs a complete and unconstrained search for the target property $\neg P$. Our approach, in contrast, leverages DeepSeq3's output to predict a high-risk state set $S_i$ for the unrolling depth $i$, which is most likely to contain a counterexample satisfying the formula $\neg P$. 
As Fig~\ref{fig:BMC} shows, the solver then executes a constrained search over $\neg P \wedge L_i$, where $L_i$ is the logical constraint derived from the predicted set $S_i$. This mechanism enables the solver to prioritize the exploration of a narrow subspace $L_i$ and effectively prune large parts of the search space irrelevant to the property violation. If this constrained search fails, the solver falls back to the original search, thereby ensuring the completeness of the method. Notably, to prevent overly precise guidance from frequently rolling back, we only predict a subset of the overall FF values.

\textbf{Data.} We test DeepSeq3-guided BMC (Guided BMC) against Raw BMC (using ABC’s default heuristics) on 40 randomly selected BMC instances from HWMCC benchmarks with the 1000s time limitation set to 1000 seconds and the maximum unrolling depth varied between 20 and 100. Of the total instances in the dataset, 31 are solvable within 10 seconds and thus regarded as easy, while 9 are categorized as hard. 

\begin{figure}[!t]
   \centering
   \includegraphics[width=0.75\linewidth]{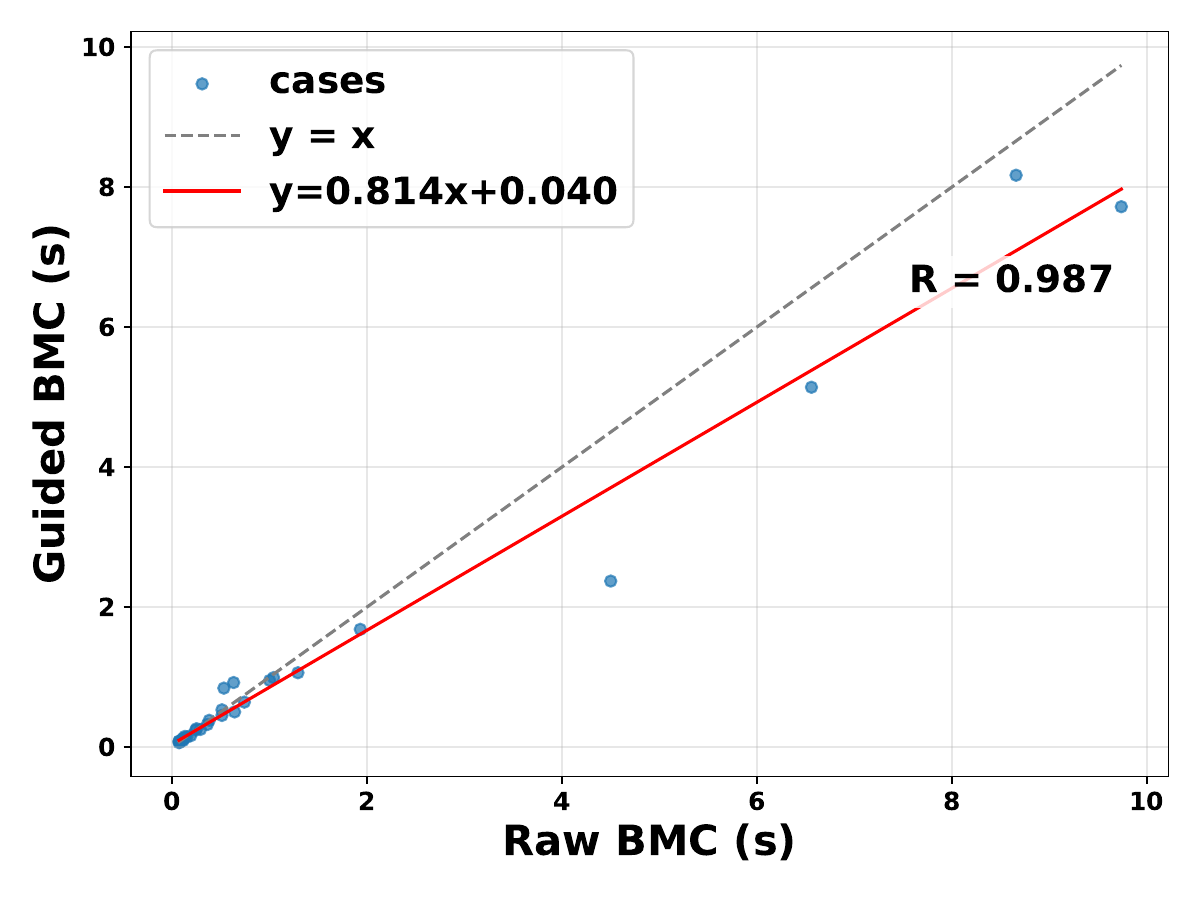} \vspace{-10pt}
   \caption{Solving time comparison on easy instances}
   \label{fig:BMC1}
   \vspace{-10pt}
\end{figure}

\textbf{Results.}
The results shows that Guided BMC surpasses Raw BMC in efficiency. 
Fig~\ref{fig:BMC1} plots the solving times for 31 easy instances. The graph excludes the model inference time to focus on the pure speedup. It achieves an average speedup of $18.6\%$ (based on the regression line $y=0.814x+0.040$). While the inclusion of model inference may increase runtime on easy cases, we do not target such trivial instances. Our interest lies in the difficult cases where BMC is slow, and for those our method delivers clear and consistent speedup. Table~\ref{tab:guided_bmc_results} lists 9 difficult cases. It revelas that Guided BMC shows superior efficiency with an average total solving time of $191.11s$ compared to Raw BMC's average of $255.03s$, and successfully solves instances that Raw BMC fails due to timeout. 

\begin{table}[t]
\setlength{\tabcolsep}{2.5pt}
\scriptsize
\centering
\caption{Solving time comparison on hard instances}
\label{tab:guided_bmc_results} \vspace{-5pt}
\resizebox{\columnwidth}{!}{
\begin{tabular}{l c c c c c c}
\toprule
\multirow{2}{*}{\textbf{Circuit}} & \multirow{2}{*}{\textbf{Nodes}} & \textbf{Raw BMC\cite{brayton2010abc}} & \multicolumn{3}{c}{\textbf{Guided BMC}} \\
\cmidrule(lr){3-6}
& & \textbf{Time (s)} & \textbf{SolveTime (s)} & \textbf{InferTime (s)} & \textbf{Total (s)} \\
\midrule
eijks208o & 172 & 15.63 & 11.97 & 0.74(0.5+0.24) & \textbf{12.71}\\
pdtvispeterson & 712 & 22.92 & 18.27 & 1.01(0.75+0.26) & \textbf{19.28}\\
pdtvisgigamax0 & 1,107 & \textbf{13.99} & 14.61 & 2.25(2.00+0.25) & 16.86\\
float\_reg\_b1\_1071 &2,066 & \textbf{11.02} & 10.93 & 1.30(1+0.30) &12.23\\
float8 & 10,590 & 71.05 & 61.37 & 7.81(7.5+0.31) & \textbf{69.18}\\
bj08amba5g62 & 20,009 & 133.43 & 130.83 & 1.25(1+0.25) & \textbf{132.8}\\
pc\_sfifo\_3+token\_ring.04 & 56,611 & Timeout(1000) & 946.40 & 6.30(6+0.30) & \textbf{952.70} \\
AllInterval-019 & 63,156 & Timeout(1000) & 458.72 & 10.12(10+0.24) & \textbf{468.64}\\
Problem04\_label27 & 95,960 & \textbf{27.21} & 20.74 & 15.33(15+0.33)& 36.07 \\
\midrule
\qquad \qquad \qquad \textbf{Avg.} & & 255.03 & 185.98& 5.13 & \textbf{191.11}\\
\bottomrule
\multicolumn{4}{l}{} \\
\end{tabular}
}\vspace{-5pt}
\end{table}
\section{Conclusion and Future Work}\label{Sec:Conclusion}
We present DeepSeq3 to address the two key challenges in sequential CRL: limited scalability and insufficient temporal modeling. By partitioning the circuit along FF boundaries into an SNG and applying state-centric pretraining, DeepSeq3 efficiently captures register-level semantics at scale. Experiments confirm its advantages, establishing a strong foundation for semantic-rich sequential analysis in industrial settings. Future work may explore using its temporal representations for tasks such as retiming, state encoding, and advanced formal verification.

\newpage

\section*{Acknowledgments}
This work was supported in part by the Hong Kong Research Grants Council (RGC) under Grant No. 14202824, 14205925, C6003-24Y, and T46-415/25-R.

\balance
\bibliographystyle{ACM-Reference-Format}
\bibliography{reference}

@String{CACM = "Communications of the {ACM}" }

@String{CACM = "Commun. {ACM}" }

@String{JACM = "J. ACM" }

@String{Computer = "{IEEE} Computer" }

@String{Academic = "Academic Press" }

@String{Springer = "Springer-Verlag" }

@Inproceedings{li2022deepgate,
  title="Deepgate: Learning neural representations of logic gates",
  author="Li, Min and Khan, Sadaf and Shi, Zhengyuan and Wang, Naixing and Yu, Huang and Xu, Qiang",
  booktitle="Proceedings of the 59th ACM/IEEE Design Automation Conference",
  pages="667--672",
  year="2022"
}

@inproceedings{brayton2010abc,
  title={ABC: An academic industrial-strength verification tool},
  author={Brayton, Robert and Mishchenko, Alan},
  booktitle={Computer Aided Verification},
  pages={24--40},
  year={2010},
  organization={Springer}
}

@inproceedings{shi2023deepgate2,
  title={DeepGate2: Functionality-Aware Circuit Representation Learning},
  author={Shi, Zhengyuan and Pan, Hongyang and Khan, Sadaf and Li, Min and Liu, Yi and Huang, Junhua and Zhen, Hui-Ling and Yuan, Mingxuan and Chu, Zhufei and Xu, Qiang},
  booktitle={Proceedings of the 2023 IEEE/ACM international conference on Computer-aided design},
  year={2023}
}

@inproceedings{shi2024deepgate3,
  title={Deepgate3: Towards scalable circuit representation learning},
  author={Shi, Zhengyuan and Zheng, Ziyang and Khan, Sadaf and Zhong, Jianyuan and Li, Min and Xu, Qiang},
  booktitle={Proceedings of the 2024 IEEE/ACM international conference on Computer-aided design},
  year={2024}
}

@article{khan2023deepseq,
  title={DeepSeq: Deep Sequential Circuit Learning},
  author={Khan, Sadaf and Shi, Zhengyuan and Li, Min and Xu, Qiang},
  journal={arXiv preprint arXiv:2302.13608},
  year={2023}
}

@inproceedings{liu24polargate,
    author = {Jiawei Liu and Jianwang Zhai and Mingyu Zhao and Zhe Lin and Bei Yu and Chuan Shi},
    title = {PolarGate: Breaking the Functionality Representation Bottleneck of And-Inverter Graph Neural Network},
    booktitle = {Proceedings of the 43rd IEEE/ACM International Conference on Computer-Aided Design},
    year = {2024}
}

@inproceedings{Khan2024DeepSeq2,
  author={Sadaf Khan and Zhengyuan Shi and Ziyang Zheng and Min Li and Qiang Xu},
  title={{DeepSeq2: Enhanced Sequential Circuit Learning with Disentangled Representations}},
  booktitle={Proceedings of the 30th Asia and South Pacific Design Automation Conference (ASP-DAC '25)},
  year={2024}
}

@article{mishchenko2007abc,
  title={ABC: A system for sequential synthesis and verification},
  author={Mishchenko, Alan and others},
  journal={URL http://www. eecs. berkeley. edu/alanmi/abc},
  volume={17},
  year={2007}
}

@inproceedings{wu2023gamora,
  title={Gamora: Graph learning based symbolic reasoning for large-scale boolean networks},
  author={Wu, Nan and Li, Yingjie and Hao, Cong and Dai, Steve and Yu, Cunxi and Xie, Yuan},
  booktitle={2023 60th ACM/IEEE Design Automation Conference (DAC)},
  pages={1--6},
  year={2023},
  organization={IEEE}
}

@inproceedings{deng2024less,
  title={Less is more: Hop-wise graph attention for scalable and generalizable learning on circuits},
  author={Deng, Chenhui and Yue, Zichao and Yu, Cunxi and Sarar, Gokce and Carey, Ryan and Jain, Rajeev and Zhang, Zhiru},
  booktitle={Proceedings of the 61st ACM/IEEE Design Automation Conference},
  pages={1--6},
  year={2024}
}

@article{shi2025deepcell,
  title={DeepCell: Multiview Representation Learning for Post-Mapping Netlists},
  author={Shi, Zhengyuan and Ma, Chengyu and Zheng, Ziyang and Zhou, Lingfeng and Pan, Hongyang and Jiang, Wentao and Yang, Fan and Yang, Xiaoyan and Chu, Zhufei and Xu, Qiang},
  journal={arXiv preprint arXiv:2502.06816},
  year={2025}
}

@inproceedings{wang2025moss,
  title={MOSS: Multi-Modal Representation Learning on Sequential Circuits},
  author={Wang, Mingjun and Sun, Bin and Mu, Jianan and Gu, Feng and Han, Boyu and Yang, Tianmeng and Zhang, Xinyu and Liu, Silin and Wen, Yihan and Wang, Hui and others},
  booktitle={2025 62nd ACM/IEEE Design Automation Conference (DAC)},
  pages={1--7},
  year={2025},
  organization={IEEE}
}

@article{zheng2025deepgate4,
  title={Deepgate4: Efficient and effective representation learning for circuit design at scale},
  author={Zheng, Ziyang and Huang, Shan and Zhong, Jianyuan and Shi, Zhengyuan and Dai, Guohao and Xu, Ningyi and Xu, Qiang},
  journal={arXiv preprint arXiv:2502.01681},
  year={2025}
}

@article{fang2025nettag,
  title={NetTAG: A Multimodal RTL-and-Layout-Aligned Netlist Foundation Model via Text-Attributed Graph},
  author={Fang, Wenji and Li, Wenkai and Liu, Shang and Lu, Yao and Zhang, Hongce and Xie, Zhiyao},
  journal={arXiv preprint arXiv:2504.09260},
  year={2025}
}

@inproceedings{shi2025logic,
  title={Logic Optimization Meets SAT: A Novel Framework for Circuit-SAT Solving},
  author={Shi, Zhengyuan and Tang, Tiebing and Zhu, Jiaying and Khan, Sadaf and Zhen, Hui-Ling and Yuan, Mingxuan and Chu, Zhufei and Xu, Qiang},
  booktitle={2025 62nd ACM/IEEE Design Automation Conference (DAC)},
  pages={1--7},
  year={2025},
  organization={IEEE}
}

@inproceedings{hwmcc24,
    author = {Preiner, Mathias and Froleyks, Nils and Biere, Armin},
    title = {Hardware Model Checking Competition 2024},
    booktitle = {Proceedings of the 24th Conference on Formal Methods in Computer-Aided Design – FMCAD 2024},
    year = {2024}
}

@misc{HWMCC24Data,
    author       = {Mathias Preiner and Nils Froleyks and Armin Biere},
    title        = {{HWMCC'24 Benchmarks and Results}},
    month        = nov,
    year         = {2024},
    publisher    = {Zenodo},
    version      = {v3},
    doi          = {10.5281/zenodo.14156844}, 
    url          = {https://zenodo.org/records/14156844}
}

@misc{HWMCC20Data,
    author = {Armin Biere and Nils Froleyks and Mathias Preiner},
    title = {{Hardware Model Checking Competition 2020 Benchmarks}},
    note = {Available from the official competition website},
    url = {https://fmv.jku.at/hwmcc20/},
    year = {2020}
}

@inproceedings{shi22deeptpi,
    author = {Zhengyuan Shi and Min Li and Sadaf Khan and Liuzheng Wang and Naixing Wang and Yu Huang},
    title = {DeepTPI: Test Point Insertion with Deep Reinforcement Learning},
    booktitle = {2022 IEEE International Test Conference (ITC)},
    page = {194–203},
    year = {2022}
}

@inproceedings{Limin23sat,
    author = {Min Li and Zhengyuan Shi and Qiuxia Lai and Sadaf Khan and Shaowei Cai and Qiang Xu.},
    title = {On eda-driven learning for sat solving},
    booktitle = { Proceedings of the 60th Annual ACM/IEEE Design Automation Conference},
    page ={1--6},
    year = {2023}
}

@article{zhu2025circuit,
  title={Circuit-Aware SAT Solving: Guiding CDCL via Conditional Probabilities},
  author={Zhu, Jiaying and Zheng, Ziyang and Shi, Zhengyuan and Cai, Yalun and Xu, Qiang},
  journal={arXiv preprint arXiv:2508.04235},
  year={2025}
}

@book{handbookbmc,
    editor = {Edmund M. Clarke and Thomas A. Henzinger and Helmut Veith and Roderick Bloem},
    title = {{Handbook of Model Checking}},
    publisher = {Springer},
    year = {2018},
    isbn = {978-3-319-10574-1},
}

@techreport{Brglez1989ISCAS,
  author = {Brglez, Franc and Bryan, David and Kozminski, Krzysztof},
  title = {Notes on the ISCAS'89 Benchmark Circuits},
  institution = {North-Carolina State University},
  year = {1989},
  type = {Technical Report}
}

@inproceedings{Davidson1999ITC,
  author = {Davidson, Scott},
  title = {Characteristics of the {ITC'99} Benchmark Circuits},
  booktitle = {Proceedings of the International Test Conference (ITSW)},
  year = {1999}
}

@online{Opencores,
  author = {{Opencores Team}},
  title = {{Opencores}},
  url = {https://opencores.org/}
}

@inproceedings{kipf2017semi,
  title={Semi-Supervised Classification with Graph Convolutional Networks},
  author={Kipf, Thomas N. and Welling, Max},
  booktitle={International Conference on Learning Representations (ICLR)},
  year={2017}
}

@inproceedings{velickovic2018gat,
  title={Graph Attention Networks},
  author={Veli{\v{c}}kovi{\'c}, Petar and Cucurull, Guillem and Casanova, Arantxa and Romero, Adriana and Li{\`o}, Pietro and Bengio, Yoshua},
  booktitle={International Conference on Learning Representations (ICLR)},
  year={2018}
}

@inproceedings{graphsage,
  title={Inductive Representation Learning on Large Graphs},
  author={Hamilton, William L. and Ying, Rex and Leskovec, Jure},
  booktitle={Advances in Neural Information Processing Systems (NeurIPS)},
  year={2017}
}

@inproceedings{grannite,
    author = {Yanqing Zhang and Haoxing Ren and Brucek Khailany},
    title = {GRANNITE: Graph Neural Network Inference for Transferable Power Estimation},
    booktitle = {2020 57th ACM/IEEE Design Automation Conference (DAC)},
    year = {2020}
}

@inproceedings{floyd,
    author = {Robert W. Floyd},
    title = {Algorithm 97: Shortest path},
    booktitle = {Communications of the ACM (CACM)},
    year = {1962},
    page = {345}
}

@inproceedings{warshall,
    author = {Stephen Warshall},
    title = {A Theorem on Boolean Matrices},
    booktitle = {Journal of the ACM (JACM)},
    year = {1962},
    page = {11-12}
}

@inproceedings{llm-bmc,
author = {Pirzada, Muhammad A. A. and Reger, Giles and Bhayat, Ahmed and Cordeiro, Lucas C.},
title = {LLM-Generated Invariants for Bounded Model Checking Without Loop Unrolling},
booktitle = {Proceedings of the 39th IEEE/ACM International Conference on Automated Software Engineering},
year = {2024}
}

@inproceedings{fang2025circuitfusion,
  title        = {CircuitFusion: Multimodal Circuit Representation Learning for Agile Chip Design},
  author       = {Wenji Fang and Shang Liu and Jing Wang and Zhiyao Xie},
  booktitle    = {International Conference on Learning Representations (ICLR)},
  year         = {2025}
}

@article{FrohnGiesl2024arXiv,
  author  = {Florian Frohn and J\"urgen Giesl},
  title   = {Integrating Loop Acceleration Into Bounded Model Checking},
  journal = {CoRR},
  volume  = {abs/2401.09973},
  year    = {2024},
  url     = {https://arxiv.org/abs/2401.09973}
}

\end{document}